\pdfoutput=1

\documentclass[11pt]{article}

\usepackage{emnlp2021}

\usepackage[T1]{fontenc}

\usepackage[utf8]{inputenc}

\usepackage{microtype}

\usepackage{times}
\usepackage{soul}
\usepackage{url}
\usepackage{caption}
\usepackage{graphicx}
\usepackage{amsmath}
\usepackage{amsthm}
\usepackage{booktabs}
\usepackage{algorithm}
\usepackage{algorithmic}
\usepackage{latexsym}
\usepackage{todonotes}
\usepackage{multirow}
\usepackage{enumitem}
\usepackage{newtxmath}
\usepackage{tabularx}
\usepackage{float}
\usepackage{adjustbox}
\usepackage{makecell}
\usepackage{subfigure}

\title{CausalBERT: Injecting Causal Knowledge Into Pre-trained Models with Minimal Supervision}

\author{Zhongyang Li, Xiao Ding, Kuo Liao, Bing Qin, Ting Liu  \\
    Research Center for Social Computing and Information Retrieval \\
    Harbin Institute of Technology, China \\
    {\{zyli, xding, kliao, qinb, tliu\}@ir.hit.edu.cn}}

\date{}
\begin{document}
\maketitle

\begin{abstract}

Recent work has shown success in incorporating pre-trained models like BERT to improve NLP systems. However, existing pre-trained models lack of causal knowledge which prevents today's NLP systems from thinking like humans. In this paper, we investigate the problem of injecting causal knowledge into pre-trained models. There are two fundamental problems: 1) how to collect various granularities of causal pairs from unstructured texts; 2) how to effectively inject causal knowledge into pre-trained models. To address these issues, we extend the idea of CausalBERT from previous studies, and conduct experiments on various datasets to evaluate its effectiveness.
In addition, we adopt a regularization-based method to preserve the already learned knowledge with an extra regularization term while injecting causal knowledge. Extensive experiments on 7 datasets, including four causal pair classification tasks, two causal QA tasks and a causal inference task, demonstrate that CausalBERT captures rich causal knowledge and outperforms all pre-trained models-based state-of-the-art methods, achieving a new causal inference benchmark. 

\end{abstract}

\section{Introduction}



\begin{figure}
    \centering
    \includegraphics[width=\columnwidth]{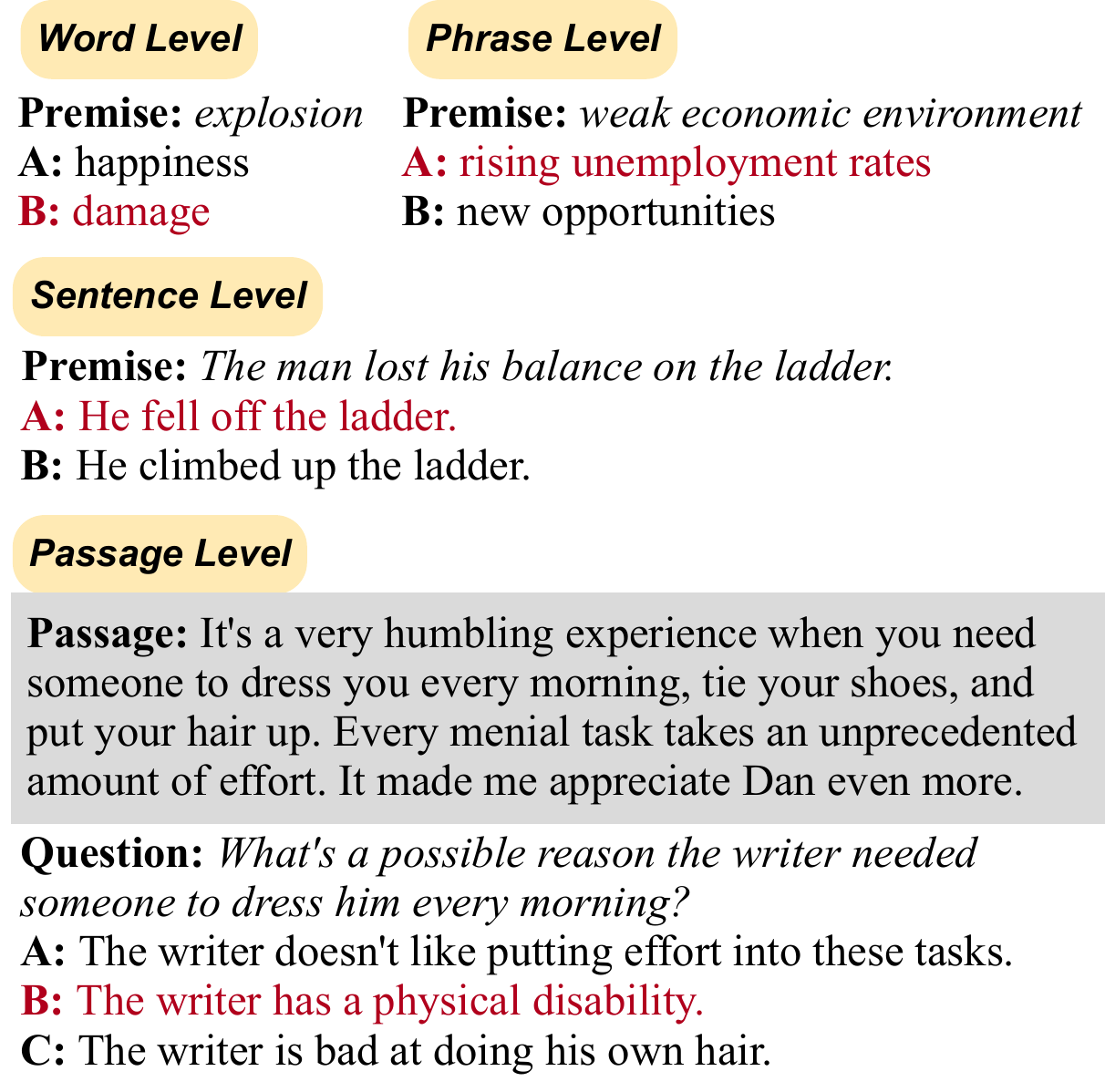}
    \vspace{-0.5cm}
    \caption{Different levels of causal inference tasks: red colored options are the correct answers.}
    \label{fig:intro}
    \vspace{-0.4cm}
\end{figure}


Pre-trained language models like GPT~\cite{radford2018improving}, BERT~\cite{devlin2018bert}, XLNet~\cite{yang2019xlnet}, and RoBERTa~\cite{liu2019roberta} have shown that a two-stage framework --- pre-training a language model on large-scale unlabeled corpora and fine-tuning on target tasks --- can bring promising improvements to various natural language understanding tasks, such as reading comprehension~\cite{radford2018improving} and natural language inference~\cite{devlin2018bert}. 

In this paper, we study a series of natural language causal inference (NLCI) problems with the form of ``\textit{Could X cause Y?}'', where \textit{X} and \textit{Y} can be single words (``explosion'', ``damage''), general phrases without any constraints (``weak economic environment'', ``rising unemployment rates''), complete sentences (``The man lost his balance on the ladder.'', ``He fell off the ladder.''), and even formulized as a complicated reading comprehension task with a long passage, a question and several candidate answers (as shown in Figure \ref{fig:intro}).

Despite the great success of the pre-trained language models in NLP systems, recent studies show that models learned in such an unsupervised manner struggle to capture causal knowledge and cannot achieve a satisfactory performance in NLCI tasks \cite{hassanzadeh2019answering,li2019learning}. Enabling pre-trained models capable of causal inference with rich causal knowledge drives us to achieve human-like AI. However, to the best of our knowledge, few studies have explored this problem, mainly due to the following two main challenges:

1. The difficulty of creating different granularities of causal pairs datasets with reasonable quality and coverage;

2. How to effectively inject causal knowledge into a unified model for solving different levels of causal inference tasks and tellingly alleviate the problem of catastrophic forgetting?


In this paper, we try to solve these two issues for better pre-trained causal inference models. Existing causal inference datasets suffer from the problem of small scale~\cite{roemmele2011choice,hassanzadeh2019answering} or low quality~\cite{Sharp2016CreatingCE,Xie2019DistributedRO}. For the dataset challenge, apart from borrowing causal knowledge from existing high-quality resources like ConceptNet~\cite{Speer2016ConceptNet5A} and CausalBank\cite{li2020guided}, we also make use of the cheap supervision from the linguistic patterns \cite{nie-etal-2019-dissent,Zhou2020TemporalCS}. Specifically, we collect several causal knowledge resources, including both the word level and sentence level cause-effect pairs, based on a series of causal patterns curated from previous studies \cite{mirza2014annotating,luo2016commonsense}. Besides, we adopt the causal word embedding techniques \cite{Xie2019DistributedRO} to automatically learn word pairs with strong causal relations. These together lead to a large-scale causal resource, which we believe can stimulate a lot of future causal inference studies.


\citet{li2020guided} illustrated that a encoder trained from a corpus of causal pair constructions (CausalBank) benefitted causal inference as measured in COPA\cite{roemmele2011choice}. This study is a downstream follow-up of the CausalBERT\cite{li2020guided} idea and investigate its effectiveness on additional tasks.  
Specifically, we devise an additional causal pair classification or ranking task for the pre-trained models based on various granularities of causal resources we collected, equipping them with causal inference abilities. Then the models can be directly applied to the causal pair classification or COPA test sets, simulating a zero-shot setting \cite{Xian2019ZeroShotLC}, or further fine-tuned on the downstream causal inference tasks. To alleviate the catastrophic forgetting~\cite{Kirkpatrick2017OvercomingCF} problem on some complicated causal QA datasets \cite{Sharp2016CreatingCE,Huang2019CosmosQM}, we adopt a regularization-based method to preserve the previous knowledge with an extra regularization term~\cite{Kirkpatrick2017OvercomingCF}.

We conduct extensive experiments on seven benchmark datasets across three causal knowledge driven tasks, i.e., causal pair classification, causal question answering and causal inference (COPA). Experiments show that CausalBERT consistently performs better than all pre-trained models-based baselines. Especially, on the well-known COPA causal inference challenge, we achieve 93.5\% accuracy with our causal knowledge enhanced ALBERT model, very close to the performance of the largest T5-11B model (94.8\%) \cite{raffel2019exploring}. Surprisingly, our CausalBERT-based ALBERT-xxlarge model achieves 86.4\% accuracy on the COPA test set under the zero-shot setting, which outperforms some strong pre-trained models fine-tuned on the COPA dataset \cite{li2019learning}. 




\section{Method}
\begin{figure}
    \centering
    \includegraphics[width=\columnwidth]{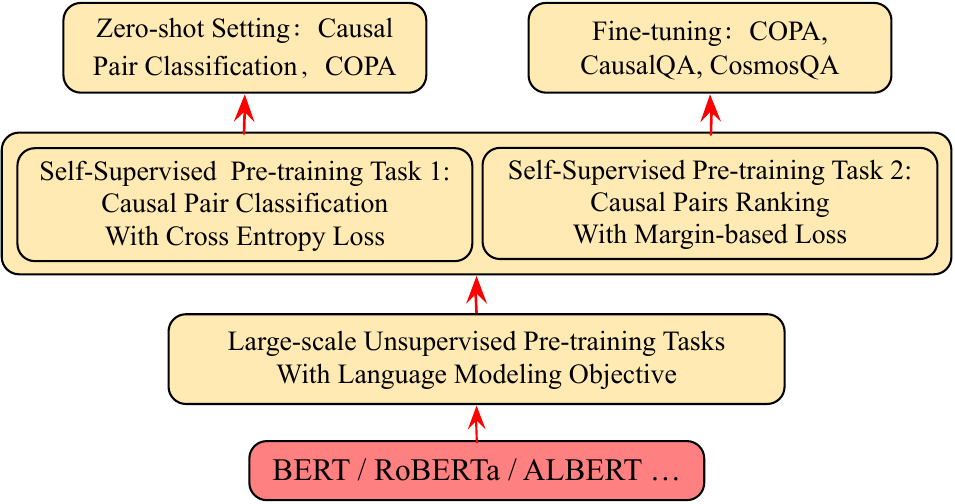}
    \vspace{-0.6cm}
    \caption{The three-stage CausalBERT framework.}
    \label{fig:method}
    \vspace{-0.4cm}
\end{figure}

As shown in Figure \ref{fig:method}, our CausalBERT is a three-stage sequential transfer learning framework \cite{ijcai2019-249,phang2018sentence}, including an unsupervised pre-training stage with the language modeling objective\footnote{We don't pre-train a language model from scratch, but use the publicly available pre-trained BERT \cite{devlin2018bert}, RoBERTa \cite{liu2019roberta} and ALBERT \cite{lan2019albert}.}, a second self-supervised pre-training stage using the different levels' causal pairs resources we collect (Section 2.1), with the proposed causal pair classification or ranking pre-training tasks (Section 2.2), and the regularization technique for overcoming catastrophic forgetting (Section 2.3). Finally, in the third stage the model can be directly applied to the causal pair classification and COPA test sets, or further fine-tuned on the target tasks' training sets.

\subsection{Where is the Causal Knowledge from?}
In order to inject causal knowledge into the pre-trained language models, we collect a large-scale and high-quality causal resource, either from previous resources or using precise causal patterns. We consider two categories of causal knowledge: sentence (phrase) level cause-effect pairs, and word level cause-effect pairs.

\subsubsection{Sentence Level Causal Knowledge}

\textbf{CausalBank.} CausalBank\cite{li2020guided} is a large-scale sentential causal pairs dataset\footnote{http://nlp.jhu.edu/causalbank} extracted from the preprocessed English Common Crawl corpus \cite{buck2014n}. It contains 314 M cause-effect pairs in total. Though very large in size, in experiment we don't use the whole corpus for causal knowledge injection but use some subsets. Future studies can decide how to use this very large causal knowledge base for tasks like causal generation \cite{rashkin2018event2mind,Sap2019ATOMICAA}, or pre-training a CausalBERT model from scratch.

\textbf{ConceptNet.}
ConceptNet \cite{Speer2016ConceptNet5A} is a knowledge graph that connects words and phrases of natural language with labeled edges. The knowledge was collected from many sources that include experts-created resources, crowd-sourcing, and games with a purpose. It uses a closed class of selected relations such as \textit{IsA, UsedFor, and CapableOf}. We obtain 22 K phrase level cause-effect pairs from the \textit{Causes} relation, such as ``A big game'' causes ``watch television''.

\subsubsection{Word Level Causal Knowledge}
As we also evaluate our method on word level causal pair classification tasks, we need to collect a word level causal pairs training set. We consider the following three approaches. 

\textbf{Cheap Supervision from Precise Template Matching.}
Inspired by \citet{Girju2002TextMF}, we propose to use some low ambiguity and precise causal patterns to extract word level cause-effect pairs from the preprocessed English Common Crawl corpus (5.14~TB) \cite{buck2014n}. Specifically, we find that some grammatical structures in English like \textit{NP1}-verb \textit{NP2}, where the verb can be `caused', `causing', `induced' and `inducing', explicitly express a causal relation between \textit{NP1} and \textit{NP2}. For example, ``\textit{virus}-caused \textit{infection}'' implies \textit{virus} causes \textit{infection}, and ``\textit{sleep}-inducing \textit{pills}'' implies \textit{pills} cause \textit{sleep}. We totally collect 558 K word level causal pairs by using this simple but effective template matching approach.


\textbf{CausalNet.} We reproduced a variant of CausalNet \cite{luo2016commonsense} in our CausalBank\cite{li2020guided} study\footnote{https://github.com/eecrazy/CausalBank}, please refer to that work for details. For experiment we keep the top 1.96 M word pairs with the highest necessity causal strength\cite{luo2016commonsense}, computed with: 
$C S_{\text {nec}}\left(i_{c}, j_{e}\right) =\frac{p\left(i_{c} | j_{e}\right)}{p^{\alpha}\left(i_{c}\right)} =\frac{p\left(i_{c}, j_{e}\right)}{p^{\alpha}\left(i_{c}\right) p\left(j_{e}\right)}$,
where $\alpha$ is a constant penalty exponent value, penalizing high-frequency response terms. $p\left(i_{c}\right)$, $p\left(j_{e}\right)$ and $p\left(i_{c}, j_{e}\right)$ are the probabilities that $i_{c}$ is a cause word in the corpus, $j_{e}$ is an effect word in the corpus, and $i_{c}$ has a causal relation with $j_{e}$.

\textbf{Causal Embedding.}
Apart from the above template matching based methods, we further adopted the causal word embedding techniques proposed by \citet{Xie2019DistributedRO} to automatically learn word pairs with strong causal relations, by running their \textit{Max-Matching} model on our sentence level CausalBank corpus. We used 100 word embedding size, with a 113 K cause word vocabulary and 92 K effect word vocabulary, running 11 epochs to reach convergence. Then we harvest the word pairs with embedding inner product similarity scores above 0, obtaining 120 K word level causal pairs.

\subsection{Pre-training Tasks For CausalBERT}
Previous studies \cite{ijcai2019-249,phang2018sentence} have shown that applying intermediate auxiliary task training to an encoder such as BERT can improve performance on a target task. In this paper, we further extend the CausalBERT\cite{li2020guided,li2019learning} idea from previous studies to conduct experiments on more datasets to evaluate its effectiveness. In order to inject causal knowledge into pre-trained language models, we devise the following two pre-training tasks to further train the models:

\textbf{(1). Causal Pair Classification with Cross Entropy Loss:} For each positive cause-effect pairs, we randomly sample some false causes or effects from other relations, to get the negative training examples. Then we pre-train the models such as BERT using a binary classification task, with a cross entropy objective.

\textbf{(2). Causal Pairs Ranking with Margin-based Loss:} Like the above classification task, we also use negative sampling to get the negative training examples. Instead of doing a binary classification, we rank the positive cause-effect pairs to have a prediction score above the negative pairs, employing a margin-based loss \cite{li2019learning,li2018constructing} in the objective function: 
$L(\Theta) = \sum_{(c,e)\in \mathcal{B}}(\max(0, m - f(c,e) + f(c',e'))) +\frac \lambda 2 ||\Theta||^2$,
where $f(c,e)$ is the score of true CE pair given by BERT model, $f(c',e')$ is the score of corrupted CE pair by replacing $c$ or $e$ with randomly sampled negative cause $c'$ or effect $e'$ from other examples. $m > 0$ is the margin loss function parameter. $\Theta$ is the set of pre-trained models' parameters. $\lambda$ is the parameter for L2 regularization. 

For causal pair classification test sets (section 3.1), we use the first pre-training task; for multiple choices test sets (section 3.2, 3.3), we use the second ranking-based pre-training task following the suggestions from \citet{li2019learning}. By training pre-trained models with these tasks, we expect the model to acquire specific domain knowledge about the meaning of a causal relation, and perform better on downstream causal inference tasks. 

In practice, for each positive cause-effect pair we randomly samples $K$ negative pairs. The label imbalance in training data may lead to a biased model. We use weight adjustment to fix this. Specifically, we apply weight adjustment to the total loss with a weight factor calculated as the observed label's count relative to the number of all instances.

\subsection{Overcoming Catastrophic Forgetting}
In order to alleviate the catastrophic forgetting issue seen in our sequential transfer learning method, we adopt a regularization-based method \cite{Chen2020RecallAL} to preserve the previous knowledge with an extra regularization term. Specifically, we apply an L2 regularization term on the pre-trained models' parameters when injecting causal knowledge into them. This simple technique helps the model don't deviate too far from the already learned language modeling parameters, while learning new causal knowledge.

\subsection{Model Details For Reproduction}
Our models have the same number of parameters as the original models in \citet{devlin2018bert,liu2019roberta,lan2019albert}. We use the following hyperparameters: 1e-5 learning rate, 8, 40, 80 or 200 batch size, 8, 21, 50, 150 or 192 max sequence length, running for at most 3 epochs, evaluating for every 50 or 300 optimization steps. Our code, running scripts with used hyperparameters and sample data of our collected causal resources are uploaded as supplementary materials. The base version of our models run for about 1 to 3 hours, while the large version run for 10 to 30 hours, according to the training data size. Each of our model uses one NVIDIA 2080ti GPU with 11GB memory. 
For each positive causal pair, we randomly samples two negative examples.

\section{Causal Inference Benchmark Datasets}

\subsection{Causal Pair Classification}

In this following we describe the four causal pair classification test sets. These datasets are created by human experts and the later three are from real-world risk management applications (https://doi.org/10.5281/zenodo.3214925). We use the four datasets for direct evaluation of our causal knowledge enhanced language models, as they are in the same form with our pre-training tasks used for injecting causal knowledge.

\textbf{SemEval.} This is the same data set used by \citet{Sharp2016CreatingCE} for the sake of comparison with state-of-the-art methods. The data set is derived from the SemEval 2010 Task 8 \cite{hendrickx2009semeval}, originally a classification of semantic relations between nominals (words). The dataset consists of 1,730 word pairs, out of which 865 (half) are marked as causal, and the rest are a random subset of non-causal relations. A positive cause-effect example is `vaccine' and `fever', while a negative pair can be `aliens' and `space'.

\textbf{NATO-SFA.} This dataset comes from a report that examines the main trends of global change and the resultant defense and security implications. This report is a result of a deep understanding of various trends and conditions throughout the world by a large number of human experts. The title text of each trend was used as a cause and the text of the implication as the effect. Random sampling was used to generate non-causal pairs, leading to totally 118 cause-effect pairs.

\textbf{Risk Models.} This causal pairs dataset was created from the models designed by expert analysts for a decision support system \cite{Sohrabi2018IBMSP,Sohrabi2018AnAP,Sohrabi2019IBMSP}. The models can be seen as graphs where nodes are short descriptions of conditions or events (e.g. High Inflation Rate or Increase in Corruption) and edge simply a causal relation. These models are based on the experts' domain knowledge in enterprise risk management. The result is a set of 804 cause-effect pairs.

\textbf{CE Pairs.} As another collection of cause-effect pairs targeting the use case in risk management, this dataset was manually extracted where either the cause or effect is from the node labels in the above risk models, but another phrase comes from online news or other documents. This dataset contains 160 causal and 160 non-causal pairs.

For comparison, we adopted state-of-the-art baselines from \citet{hassanzadeh2019answering}: the word occurrences-based \textbf{PMI}; \textbf{CEA}, a modification of the PMI that multiplies other factors; \textbf{DCC}, a search based method on a large corpus; \textbf{DCC-embed}, a word embedding based method that treats each phrase as a word; \textbf{NLM-BERT} uses BERT~\cite{devlin2018bert} to encode sentences ``X may cause Y'' and its top k most similar causal sentences, and compute the average cosine similarity score for making a prediction. We also report precision, recall, F1 and accuracy.

\subsection{Causal Inference}
\textbf{COPA.} (Choice of Plausible Alternatives, \cite{roemmele2011choice}) is a causal inference task in which a system is given a premise sentence and must determine either the cause or effect of the premise from two possible choices. All examples are handcrafted. It has 500 examples in the training set and 500 in the test set. Following the original work, we evaluate using accuracy. Balanced COPA (\textbf{BCOPA}) \cite{Kavumba2019WhenCP} extended COPA with one additional mirrored instance for each original training instance to overcome the superficial cues in COPA, that may be exploited by models like BERT. This mirrored instance uses the same alternatives as the corresponding original instance, but introduces a new premise which matches the wrong alternative of the original instance, leading to another 500 training examples.

\subsection{Causal Question Answering}
\textbf{CausalQA.} This dataset contains 3,031 causal questions extracted from the Yahoo! Answers corpus \cite{Sharp2016CreatingCE}, split into 60\%, 20\% and 20\% for train, dev and test. The questions are with simple surface patterns such as ``What causes ...'' and ``What is the result of ...''. All the answers are generated by the community, and one of them is voted as the top answer. Each causal question has 7.7 candidate answers on average, ranging from 4 to 93. The task is to identify the top answer from the alternatives. 

\textbf{CosmosQA.} CosmosQA \cite{Huang2019CosmosQM} is a large-scale dataset of 35,600 problems that require causal commonsense-based reading comprehension, formulated as multiple-choice questions. It focuses on reading between the lines over a diverse collection of people's everyday narratives, asking questions like ``what might be the possible reason of ...?'', or ``what would have happened if ...'' that require causal inference beyond the exact text spans in the context.

\begin{table*}[!h]\footnotesize
	\setlength\tabcolsep{1.5pt}
    \centering
    \begin{tabular}{l|cccc|cccc|cccc|cccc} 
         \toprule
         \textbf{Method} & \textbf{P} & \textbf{R} & \textbf{F1} & \textbf{Acc} & \textbf{P} & \textbf{R} & \textbf{F1} & \textbf{Acc}& \textbf{P} & \textbf{R} & \textbf{F1} & \textbf{Acc}& \textbf{P} & \textbf{R} & \textbf{F1} & \textbf{Acc}\\
         \midrule
         &\multicolumn{4}{c}{\textbf{SemEval}}&\multicolumn{4}{c}{\textbf{Risk Models}}&\multicolumn{4}{c}{\textbf{NATO-SFA}}&\multicolumn{4}{c}{\textbf{CE Pairs}}\\
         \multirow{1}{*}{PMI}&50.0 & 100.0 & 66.6 & 52.9 &50.0 & 100.0 & 66.7 & 53.1         &50.0 & 100.0 & 66.7 & 60.2 &50.0 & 100.0 & 66.6 & 50.9\\
         \multirow{1}{*}{CEA}&50.0 & 100.0 & 66.6 & 54.0& 50.0 & 100.0 & 66.7 & 54.3         &50.0 & 100.0 & 66.7 & 55.1&50.0 & 100.0 & 66.7 & 54.1\\
		 \multirow{1}{*}{DCC}&72.1 & 68.1 & 70.0 & 72.0& 50.0 & 100.0 & 66.7 & 50.8		 &50.0 & 100.0 & 66.7 & 66.1&50.0 & 100.0 & 66.7 & 55.9\\
		 \multirow{1}{*}{DCC-embed}&66.5 & 77.9 & 71.8 & 73.4&50.0 & 100.0 & 66.7 & 52.0		 &50.0 & 100.0 & 66.7 & 66.9&50.0 & 100.0 & 66.7 & 57.5\\
		 \multirow{1}{*}{NLM-BERT}&50.2 & 99.8 & 66.8 & 61.9&52.0 & 93.8 & 66.9 & 55.7		 &52.3 & 98.3 & 68.2 & 55.9&50.5 & 100.0 & 67.1 & 56.3\\
		 \midrule
 		 \multirow{1}{*}{RoBERTa-base +1}&50.0& 100& 66.7& 55.4&68.3 & 84.6 & 75.6 & 73.0		 &66.2& 83.1& 73.7& 72.0&65.1& 93.1& 76.6& 74.1\\
 		 \multirow{1}{*}{RoBERTa-base +13}&50.1& 99.9& 66.8& 55.5&70.1& 83.3& 76.1& 73.9  		 &68.5& 84.7& 75.8& 72.9&64.2& 93.1& 76.0& 70.6\\
 		 \multirow{1}{*}{RoBERTa-base +4}&81.9& 83.0& 82.4& 82.6&52.4& 92.5& 66.9& 60.8 		 &50.0& 100.0& 66.7& 61.9&53.6& 91.9& 67.7& 62.5\\
 		 \multirow{1}{*}{RoBERTa-base +45}&63.9& 76.2& 69.5& 67.0&53.4& 92.3& 67.6& 62.6 		 &52.8 &94.9 &67.9 &63.6&56.4& 85.6& 68.0& 65.0\\
 		 \multirow{1}{*}{RoBERTa-base +46}&76.5& 80.3& 78.4& 77.8&53.2& 91.0& 67.2& 61.8  		 &52.9 &93.2 &67.5 &65.3&50.0& 100.0& 66.7& 65.0\\
 		 \multirow{1}{*}{RoBERTa-base +14}&83.0&82.1&82.6&82.7&67.4& 88.8& 76.6& 73.5 		 &63.5&91.5&75.0&72.0&66.4& 87.5& 75.5& 71.6\\
 		 \multirow{1}{*}{RoBERTa-base +134}&84.8& 82.3& 83.5& 83.8&67.2 &86.8 &75.8 &72.9 		 &70.0& 83.1& 76.0& 73.7&67.8& 85.6& 75.7& 72.5\\
 		 \multirow{1}{*}{RoBERTa-base +245}&64.2 &78.3 & 70.6&67.4&69.5 & 86.6 & 77.1 & 74.3 		 &70.6&81.4&75.6&73.7&69.9&85.6&77.0&74.4\\
		 \multirow{1}{*}{RoBERTa-base +234}&83.3 & 85.1 & \textbf{84.2} & 84.0&68.0 & 87.6 & 76.5 & 74.1		 &67.5 & 88.1 & 76.5 & 74.6&69.8 & 88.1 & 77.9 & 75.0\\
		 \multirow{1}{*}{BERT-large +134}&84.6& 82.2& 83.4& 83.7& 62.1& 88.6& 73.0& 69.7		 &67.5& 88.1& 76.5& 74.6 & 63.8& 85.0& 72.9& 68.4\\
		 \multirow{1}{*}{RoBERTa-large +134}&86.9& 80.8& 83.8& 84.3&69.5& 88.8& 77.9& 76.1 		 &65.5& 93.2& 76.9& 72.9 &73.7& 87.5& 80.0& \textbf{78.4} \\
		 \multirow{1}{*}{RoBERTa-large +234}&83.1& 81.3& 82.2& 83.0&69.6& 87.8& 77.7& 75.6 		 &71.1& 91.5& 80.0& 77.1 &71.2& 91.2& 80.0& 77.2 \\
		 \multirow{1}{*}{ALBERT-xxlarge +134}&88.8& 77.3& 82.7& 83.8& 72.1& 86.8& 78.8& \textbf{77.2}		 & 76.8& 89.8& 82.8& 81.4 & 74.9& 85.6& 79.9& \textbf{78.4}\\
		 \multirow{1}{*}{ALBERT-xxlarge +234}&85.8& 82.0& 83.9& \textbf{84.4}& 72.2& 88.3& \textbf{79.4}& 77.1		 & 76.4& 93.2& \textbf{84.0}& \textbf{82.2} & 72.4& 90.0& \textbf{80.2}& 77.8\\
         \bottomrule 
    \end{tabular}
    \caption{Direct evaluation results (accuracy, \%) on the four human experts created causal pair classification datasets: the first SemEval is word level test set, while the other three are phrase or sentence level test sets. Top: baseline methods from \citet{hassanzadeh2019answering}. Bottom: CausalBERT-based approaches with various pre-trained models and knowledge source combinations. The meaning of numbers in method description: 1: 0.1 M CausalBank; 2: 1 M CausalBank; 3: 22 K ConceptNet; 4: 558 K Precise Template Matching; 5: 2 M CausalNet; 6: 120 K Causal Embedding. All numbers are size of positive cause-effect pairs from the corresponding dataset.}
    \label{tab:four_datasets}
    \vspace{-0.4cm}
\end{table*}

\begin{figure}
    \centering
    \includegraphics[width=\columnwidth]{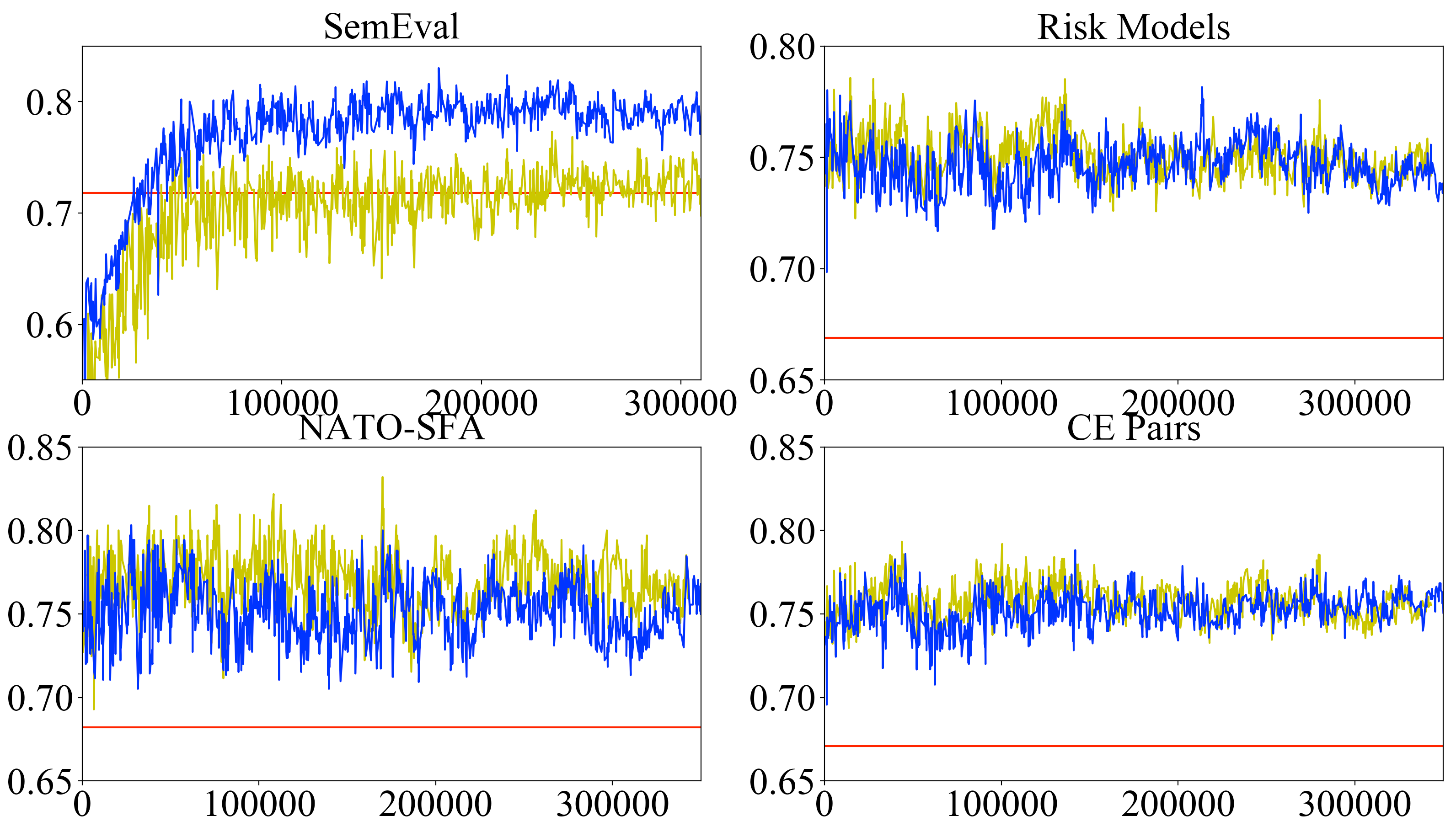}
    \vspace{-0.2cm}
	\caption{Training curves for F1 score with the number of training steps, from ALBERT-xxlarge + 234 model. Red: max baseline F1 score. Blue: with label weight adjustment. Yellow: no label weight adjustment.}
	\label{tab:training_curve}
    \vspace{-0.3cm}
\end{figure}

\section{Results and Analysis}
\textbf{Results for Causal Pair Classification.} Table \ref{tab:four_datasets} shows the direct evaluation results on the four human experts created causal pair classification datasets. We conducted extensive experiments with various pre-trained models and knowledge source combinations to study their individual impacts, using the causal pair classification pre-training task. For fair comparison with the unsupervised baselines from \citet{hassanzadeh2019answering}, we actually use the four datasets as development sets, monitoring the performance on them during training. The meaning of numbers in method description is clarified in the caption of Table \ref{tab:four_datasets}. 

By comparing the results from different methods, we can get the following observations: (1) Our methods consistently outperform the best baseline results on the four test sets, with large F1 and accuracy improvements, ranging from 10.9 to 21.5 absolute points. This demonstrates great power of CausalBERT in causal inference. (2) Word level causal knowledge from Precise Template Matching are very effective for word level SemEval test set (82.4\% F1). Adding more sentence level causal knowledge from CausalBank and ConceptNet further improve the results (83.5\% and 84.2\% F1). However, causal knowledge from CausalNet and Causal Embedding hurts the performance on SemEval, mainly due to the noise introduced in them. (3) Clean causal knowledge from CausalBank, ConceptNet and Precise Template Matching are most useful for causal pair classification, leading to the best F1 performances (79.4\%, 84.0\%, 80.2\%) on the later three sentence level tasks with ALBERT-xxlarge model \cite{lan2019albert} (RoBERTa-base \cite{liu2019roberta} gets the best F1 vaule (84.2\%) on SemEval). (4) We achieved a unified strong classifier (Causal ALBERT-xxlarge) for word, phrase and sentence level causal pair classification. 

Pre-trained models like BERT and RoBERTa are generally considered as sentence encoders to represent the meanings of relatively longer sentences. Our results on SemEval show that pre-trained models are also good at representing the meanings of short word pairs, which doesn't suffer from the superficial cues issues \cite{Kavumba2019WhenCP,gururangan-etal-2018-annotation,mccoy-etal-2019-right}. Figure \ref{tab:training_curve} shows training curves for F1 value with our causal ALBERT-xxlarge model. We find label weight adjustment is very crucial for the word level SemEval. 

\begin{table}[tp]\footnotesize
	\setlength\tabcolsep{2.0pt}
    \centering
    \begin{tabular}{lcc} 
         \toprule
         \textbf{Method} & \textbf{COPA} & \textbf{BCOPA}\\
         \midrule
		BERT-large \cite{sap2019socialiqa}  & 75.0 & - \\
		BERT-base \cite{li2019learning}  & 75.4 & - \\
		BERT-large \cite{Kavumba2019WhenCP}  & 76.5 & 74.5 \\
		RoBERTa-large \cite{Kavumba2019WhenCP}  & 87.7 & 89.0 \\
		RoBERTa-large (Leaderboard)  & 90.6 & - \\
		\midrule
		BERT-base (our imple.) & 74.5 & 76.3 \\
		BERT-large (our imple.)& 77.8 & 80.0 \\
		RoBERTa-base (our imple.)& 80.5 & 81.3 \\
		RoBERTa-large (our imple.)& 90.3 & 90.2 \\
		ALBERT-large (our imple.)& 80.1 & 79.7 \\
		ALBERT-xxlarge (our imple.) & 92.1 & 92.3 \\
		BERT-base + CB (0.1 M) & 78.6 & 78.6 \\
		BERT-large + CB (0.1 M) & 79.3 & 80.6 \\
		RoBERTa-base + CB (0.1 M) & 85.4 & 83.8 \\
		RoBERTa-large + CB (0.1 M) & 90.9 & 90.5 \\
		ALBERT-large + CB (0.1 M)& 82.1 & 81.5 \\
		ALBERT-xxlarge + CB (0.1 M)& \textbf{92.6} & \textbf{93.5} \\
         \bottomrule 
    \end{tabular}
    \caption{Accuracy (\%) results on COPA and BCOPA test set: fine-tuning the whole model.}
    \label{tab:copa_result_fine_tuning_whole}
    \vspace{-0.4cm}
\end{table}

\textbf{Results on COPA and BCOPA.} Table \ref{tab:copa_result_fine_tuning_whole} and Table \ref{tab:copa_result_zero_shot} show the accuracy results on COPA test set (BCOPA use the same test set) under two different settings: fine-tuning the whole model including the pre-trained models' and the output layer' parameters on the training set, pre-training the models on 0.1 M CausalBank (CB) without fine-tuning on the COPA training set (zero-shot setting). Compared with the current SOTA models, our causal knowledge enhanced models consistently achieve better results. We achieve 93.5\% accuracy with the causal knowledge enhanced ALBERT-xxlarge model, very close to the performance of the largest google-T5-11B model (94.8\%) \cite{raffel2019exploring}. Surprisingly, our causal knowledge enhanced ALBERT-xxlarge model achieves 86.4\% accuracy on the COPA test set under the zero-shot setting (Table \ref{tab:copa_result_zero_shot}), which outperforms some strong pre-trained models fine-tuned on the COPA dataset \cite{li2019learning}. This implies that CausalBank contains rich causal knowledge.

\begin{table}[tp]\small  
	\setlength\tabcolsep{2.0pt}
    \centering
    \begin{tabular}{lcc} 
         \toprule
         \textbf{Method} & \textbf{COPA}\\
         \midrule
		BigramPMI \cite{Goodwin2012UTDHLTCS} & 63.4 \\
		PMI \cite{gordon2011commonsense} & 65.4 \\
		CausalNet + PMI \cite{luo2016commonsense} & 70.2 \\
		Multiword + PMI \cite{Sasaki2017HandlingME} & 71.4 \\
		\midrule
		BERT-base + CB (0.1 M) & 67.8 \\
		BERT-large + CB (0.1 M) & 70.2 \\
		RoBERTa-base  + CB (0.1 M) & 74.0 \\
		RoBERTa-large + CB  & 77.8 \\
		RoBERTa-large + CB (0.1 M) & 82.2 \\
		ALBERT-base + CB  & 62.0 \\
		ALBERT-large + CB (0.1 M)& 68.2 \\
		ALBERT-xlarge + CB  & 72.4 \\
		ALBERT-xxlarge + CB (0.1 M) & \textbf{86.4} \\
         \bottomrule 
    \end{tabular}
    \caption{Accuracy (\%) on COPA: zero-shot setting.}
    \label{tab:copa_result_zero_shot}
    \vspace{-0.2cm}
\end{table}

\textbf{Results on CausalQA.} Table \ref{tab:cqa_results} shows the results on CausalQA. With the causal knowledge from CausalBank (0.1M CB), BERT-large \cite{devlin2018bert} improves accuracy from 38.5\% to 39.1\%, while RoBERTa-large shows decreased performance from 39.5\% to 37.8\%. We don't observe very large performance improvements on the CausalQA task like in the previous experiments. We guess this is because CausalQA dataset comes from online QA forum, with many user-generated unregular contents. And the task format is more complicated with 7.7 candidate answers on average (93 at most). On the other hand, the model may suffer from catastrophic forgetting \cite{Kirkpatrick2017OvercomingCF} in our sequential transfer learning process. We adopted two kinds of techniques to alleviate this: an L2 regularization based approach, and a parameter isolation-based method (KA, \citet{Wang2020KAdapterIK}). Results show that they are both effective in overcoming catastrophic forgetting, leading to consistent performance improvements. Finally, we get the best accuracy of 40.1\% with CausalBERT and K-adapter \cite{Wang2020KAdapterIK}.

\begin{table}[tp]\small  
    \centering
    \begin{tabular}{lc} 
         \toprule
          \textbf{Method} & \textbf{CausalQA}\\
         \midrule
		Causal Embedding \cite{Sharp2016CreatingCE} & 37.3 \\ 
		Causal Embedding \cite{Xie2019DistributedRO} & 37.9 \\
		\midrule
		BERT-large (our imple.) & 38.5 \\
		RoBERTa-large (our imple.) & 39.5 \\
		BERT-large + CB (0.1 M)& 39.1 \\
		RoBERTa-large + CB (0.1 M)& 37.8 \\
		BERT-large + L2 + CB (0.1 M, L2=0.01) & 39.5 \\
		RoBERTa-large +L2 + CB (L2=0.1) & 39.1 \\
		RoBERTa-large + L2 + CB (0.1 M, L2=0.01) & 39.8 \\
		BERT-base+concat + CB & 37.2 \\
		RoBERTa-base +concat + CB & 36.8 \\
		RoBERTa-large +KA + CB (0.1 M)& 36.8 \\
		BERT-large +KA + CB (0.1 M)& \textbf{40.1} \\
       \bottomrule 
    \end{tabular}
    \caption{Accuracy results (\%) on CausalQA test set.}
    \label{tab:cqa_results}
    \vspace{-0.4cm}
\end{table}

\textbf{Results on CosmosQA.} Table \ref{tab:cosmosqa_results} (middle) shows the results on CosmosQA development set (Labels are not released for the test set examples), which is a more complicated causal question answering task, with a long passage context in each example. From Table \ref{tab:cosmosqa_results} we can get similar observations to CausalQA that the improvements are rather minimal after injecting causal knowledge into the pre-trained models. With more causal pairs from CausalBank (1M CB), CausalBERT cannot always get better results. Even with the K-adapter (KA, \cite{Wang2020KAdapterIK}) based technique, our methods don't get significant improvements. We guess this is because CosmosQA requires the system to first understand the passage and the question, and then choose the correct answer based on indirect causal inference (not a direct causal relation classification). Thus, the relatively simple causal pair classification task brings very little benefits for this complicated causal inference task. We leave it as future work to explore more effective causal knowledge injection approach for CosmosQA. The good thing is that our implementation of the ALBERT-xxlarge model gets very high accuracy of 85.0\%, then to 85.8\% with causal knowledge injection. This outperforms the previous SOTA result of 81.8\% with a large margin. We further conduct experiments on CosmosQA without the passage information. Results are shown at the bottom of Table \ref{tab:cosmosqa_results}. We find that our CausalBERT consistently outperforms the original pre-trained models, demonstrating the effectiveness of our method.

\begin{table}\footnotesize
	\setlength\tabcolsep{0.1pt}
    \centering
    \begin{tabular}{lcc} 
         \toprule
         \textbf{Method} & \textbf{CosmosQA}\\
         \midrule
		BERT-large \cite{Huang2019CosmosQM} & 66.2 \\
		BERT-large +SWAG \cite{Huang2019CosmosQM} & 67.8 \\
		BERT-large Multiway \cite{Huang2019CosmosQM} & 68.3 \\
		RoBERTa-large \cite{Wang2020KAdapterIK} & 80.6 \\
		RoBERTa-large +MT \cite{Wang2020KAdapterIK} & 81.2 \\
		RoBERTa-large +KA \cite{Wang2020KAdapterIK} & 81.8 \\
		\midrule
		BERT-large (our imple.) & 66.5 \\
		RoBERTa-large (our imple.) & 80.7 \\
		ALBERT-xxlarge (our imple.) & 85.0 \\
		BERT-large + CB (0.1 M)& 67.6 \\
		RoBERTa-large + CB (0.1 M)& 81.0 \\
		BERT-large + KA + CB (0.1 M)& 65.5 \\
		BERT-large + KA + CB (1M) & 66.0 \\
		RoBERTa-large + KA + CB (0.1 M)& 80.8 \\
		RoBERTa-large + KA + CB (1M) & 80.1 \\
		ALBERT-xxlarge + CB (0.1 M)& \textbf{85.8} \\
		\midrule
		BERT-large (our imple.) & 58.7 \\
		RoBERTa-large (our imple.)& 63.1 \\
		BERT-large + CB (0.1 M)& 59.1 \\
		RoBERTa-large + CB (0.1 M)& 64.9 \\
         \bottomrule 
    \end{tabular}
    \caption{Accuracy results (\%) on CosmosQA dev set.}
    \label{tab:cosmosqa_results}
    \vspace{-0.3cm}
\end{table}


\section{Related Work}
\textbf{Injecting Domain Knowledge Into Pre-trained Models.}
This paper relates to the studies of injecting specific domain knowledge into pretrained language models such as BERT. TacoLM~\cite{Zhou2020TemporalCS} proposes exploiting explicit and implicit mentions of temporal common sense extracted from a large corpus to build a temporal common sense language model.
GenBERT~\cite{Geva2020InjectingNR} injects numerical reasoning skills into pre-trained language models by generating large amounts of numerical and textual data, and training in a multi-task setup.
ERNIE~\cite{Zhang2019ERNIEEL} injects a knowledge graph into BERT by aligning entities from Wikipedia sentences to fact triples in WikiData. 
SenseBERT~\cite{Levine2019SenseBERTDS} injects word-supersense knowledge by predicting the WordNet supersense of the masked word in the input.
KnowBERT~\cite{Peters2019KnowledgeEC} incorporates knowledge bases into BERT using Knowledge attention and recontextualization, where the knowledge comes from synset-synset and lemma-lemma relationships in WordNet.
In this paper, we propose to inject causal knowledge into pre-trained models, which none of previous work has explored.

\textbf{Causal Reasoning Tasks in NLP.}
COPA \cite{roemmele2011choice} is a causal inference task in which a system is given a premise and must determine either the cause or effect of the premise from two candidate answers.
\cite{Sharp2016CreatingCE} was the first work to train causal word embeddings for causal question answering.
\citet{hassanzadeh2019answering} investigated a series of unsupervised methods for answering binary causal questions. \citet{Xie2019DistributedRO} proposed three causal word embedding models which can map the labels of sentence level cause-effect pairs to word level causal relations. Then the learned causal word embeddings were used in word pairs classification and causal question answering.
\citet{Huang2019CosmosQM} proposed a machine reading comprehension style causal commonsense reasoning dataset.
In this work we gather most of the current causal inference datasets to form a novel causal inference benchmark.


\section{Conclusion}
In this paper, we extend the idea of CausalBERT\cite{li2020guided} from previous studies, and conduct experiments on various datasets to evaluate its effectiveness. Extensive experiments show that CausalBERT is effective in solving different text level's causal inference tasks, and achieves new SOTA or comparable results on seven causal inference benchmark datasets. 

\newpage
\bibliographystyle{acl_natbib}
\bibliography{emnlp2021}
\end{document}